\documentclass[letterpaper]{article} 
\usepackage{aaai2026}

\nocopyright

\usepackage{times}  
\usepackage{helvet}  
\usepackage{courier}  
\usepackage[hyphens]{url}  
\usepackage{graphicx} 
\usepackage{natbib}  
\usepackage{caption} 
\usepackage{color}
\usepackage{algorithm}
\usepackage{algorithmic}

\usepackage{newfloat}
\usepackage{listings}
\DeclareCaptionStyle{ruled}{labelfont=normalfont,labelsep=colon,strut=off} 
\floatstyle{ruled}
\newfloat{listing}{tb}{lst}{}
\floatname{listing}{Listing}
\title{``Are We Done Yet?": A Vision-Based Judge for Autonomous Task Completion of Computer Use Agents}
\title{``Are We Done Yet?": A Vision-Based Judge for Autonomous Task Completion of Computer Use Agents}

\author{
    Marta Sumyk\textsuperscript{\rm 1},
    Oleksandr Kosovan\textsuperscript{\rm 1},
}
\affiliations{
    \textsuperscript{\rm 1}Ukrainian Catholic University\\
    sumyk.pn@ucu.edu.ua, o.kosovan@ucu.edu.ua
}

\begin{document}

\maketitle
 
\begingroup
\renewcommand\thefootnote{}
\footnotetext{This work has been accepted to appear at the AAAI 2026 Workshop on Trust and Control in Agentic AI (TrustAgent).}
\addtocounter{footnote}{1} 
\endgroup

\begin{abstract}
Computer Use Agents (CUAs) are designed to autonomously operate digital interfaces, yet they often fail to reliably determine whether a given task has been successfully completed. We present an autonomous evaluation and feedback framework that leverages Vision–Language Models (VLMs) to assess task completion directly from screenshots and task descriptions. Our dataset covers $42$ built-in macOS applications and $1{,}260$ human-labeled tasks, covering a wide range of scenarios. Our framework achieves up to $73\%$ classification accuracy in task success detection and yields an average relative improvement of $27\%$ in the overall task success rate of CUAs when evaluator feedback is applied. These results demonstrate that vision-based evaluation can serve as an actionable feedback mechanism that significantly improves the reliability and self-correction of autonomous computer-use agents.

\end{abstract}\


\section{Introduction}

In recent years, Computer Use Agents (CUAs) \cite{saunders2022self, openai2025cua, litecua2025} have emerged as a promising paradigm for enabling AI systems to autonomously interact with digital environments, perceiving screen states and performing actions such as clicking, typing, and executing commands to accomplish user-specified goals. Despite their generality and service-agnostic design, a key limitation remains: CUAs often struggle to reliably determine whether a task has been successfully completed. This shortcoming manifests in two critical ways:

\begin{itemize}
\item The agent declares a task complete when it is not, undermining user trust and overall reliability \cite{sun2025seagentselfevolvingcomputeruse, sager2025comprehensivesurveyagentscomputer}.
\item The agent successfully completes the task but fails to recognize this, leading to redundant actions and unnecessary computational overhead \cite{sager2025comprehensivesurveyagentscomputer}.
\end{itemize}

To address these challenges, this work proposes a method for autonomous evaluation of task completion for macOS CUAs, aimed at improving both the success rate and the reliability of CUAs.

We focus on the macOS environment for two main reasons. First, it remains an underexplored domain in the study of screen representations and CUAs \cite{screen2ax}. While prior work has primarily examined web and mobile environments \cite{pan2024autonomousevaluationrefinementdigital, li2024seeact, humphreys2024webarena}, desktop operating systems have received considerably less attention. Desktop interfaces are also inherently more complex to interpret: they typically contain a larger number of visual elements than mobile UIs and lack structured unified representations such as Hypertext Markup Language (HTML) trees available in web environments \cite{screen2ax}. Second, we choose macOS as an ideal starting point because it offers a controlled yet diverse collection of built-in applications that enable systematic benchmarking of task execution and evaluation. In the future, we aim to extend our framework to other desktop operating systems, including Windows, Linux, and cross-platform web interfaces, toward developing general-purpose and reliable CUAs.

To sum up, the contributions of this work are as follows:

\begin{itemize}
    \item We introduce a diverse, human-labeled dataset comprising $1{,}260$ tasks across $42$ built-in macOS applications.  
    \item We propose a methodology for autonomous evaluation of task completion, achieving up to $73\%$ accuracy and improving success rate of CUAs by an average of $27\%$ relative percentage points on average. 
\end{itemize}

\section{Related Works}

\subsection{Computer Use Agents}

CUAs are autonomous agents designed to interact directly with Graphical User Interfaces (GUIs), performing actions such as clicking, typing, scrolling, dragging or navigating web pages. Early examples include browser-based assistants and recent prototypes such as OpenAI's Computer Use tool\footnote{\url{https://openai.com/index/computer-using-agent/}}, which integrate vision-language reasoning with low-level action execution. Similar lines of research explore autonomous UI navigation \cite{gur2023browsergym}, multi-modal planning \cite{li2024seeact}, and end-to-end web automation benchmarks \cite{humphreys2024webarena}.

Unlike API-based or function-calling agents that require explicit integration with each service, CUAs operate in a service-agnostic manner: they perceive and act directly on the screen, enabling interaction with any digital environment without additional engineering effort. This design paradigm makes CUAs inherently more flexible and scalable, capable of generalizing across software systems and interfaces \cite{sun2025seagentselfevolvingcomputeruse, sager2025comprehensivesurveyagentscomputer}. However, this generality also introduces new challenges in reasoning and verification. Because CUAs rely solely on visual observations, they can fail silently or partially when confronted with unexpected interface states, visual occlusions, or distribution shifts \cite{gur2023browsergym, humphreys2024webarena, li2024seeact}.
 This highlights the need for reliable mechanisms to assess whether the intended task has actually been completed, especially when the task cannot be trivially reduced to log-based success signals.

\subsection{Autonomous Evaluation of Task Completion}

A persistent challenge across all types of agentic systems is evaluating whether a goal has truly been achieved \cite{zhou2025autoevalautonomousevaluationgeneralist, bhonsle2025autoevaljudgegeneralagentic, zhuge2024agentasajudgeevaluateagentsagents}.
Reliable evaluation is fundamental for measuring performance, enabling self-improvement, and establishing user trust of agents.

In this work, we adapt this broader evaluation challenge to the domain of CUAs. The problem of determining whether a CUA has successfully completed a task has not been extensively explored \cite{sager2025comprehensivesurveyagentscomputer}. While prior work has focused primarily on improving action planning and interface understanding \cite{gur2023browsergym, li2024seeact, humphreys2024webarena}, it has paid little attention to how to tell when a task is actually completed.

A notable exception is script-based evaluation, used in the OSWorld benchmark \cite{xie2024osworldbenchmarkingmultimodalagents}. However, this approach relies on manually written verification scripts for each task, which severely limits scalability and makes real-time evaluation impractical. Maintaining reliable automated evaluation across hundreds of GUI tasks requires substantial manual effort, since even minor interface or environment changes can break the scripts and invalidate results.

Also, a closely related effort is the work of \citet{pan2024autonomousevaluationrefinementdigital}, which proposes an autonomous evaluation and refinement framework for digital agents. Their method focuses on automatically assessing and improving web-based and simulated agents by reasoning over structured representations of page elements and textual feedback. Although their approach demonstrates that model-based evaluators can significantly accelerate agent learning, it operates primarily in browser and synthetic environments where the interface semantics and success states are explicitly defined. In contrast, our setting involves real desktop interfaces, specifically macOS, where the screens are harder to parse due to variety and number of elements and also have no universal way of representation as in the web HTML \cite{screen2ax}. This makes our work a complementary extension of autonomous evaluation to unstructured, multimodal environments that better mirror real-world computer use.

In contrast, the question of task completion by an agent has been more systematically studied in other domains, particularly in robotics. In robotics, recent work such as AutoEval \cite{zhou2025autoevalautonomousevaluationgeneralist} introduces autonomous evaluation frameworks for manipulation policies, reducing the reliance on human annotators or scripted success detectors. It reaches both high agreement with human annotations and reduces the human annotation time by $99\%$.

Our work builds on this line of research but adapts it to the domain of CUAs. Unlike physical robotics tasks, task completion in digital environments often lacks a straightforward ground-truth signal: for instance, whether “sending an email” was completed correctly may not be directly observable from logs alone. Inspired by AutoEval, we propose to use vision-language models as evaluators that judge whether the current desktop state corresponds to the intended outcome, providing CUAs with reliable, automated feedback.

\section{Methodology}

\subsection{Dataset}
Our dataset\footnote{\url{https://zenodo.org/records/17696742}} covers $42$ built-in macOS applications, spanning functionality productivity, communication, multimedia, system utilities, and developer tools. For each application, we define $30$ concrete tasks, resulting in a total of $1{,}260$ tasks (in comparison, the OSWorld \cite{xie2024osworldbenchmarkingmultimodalagents} constains $369$ tasks). The task set is deliberately diverse, ranging from simple actions (e.g., \textit{“Open Calendar app”}) to more complex, multi-step interactions (e.g., \textit{“Filter apps by free in App Store and open the first result”}).

This design ensures coverage across varying levels of difficulty and interaction types, allowing us to evaluate CUAs both on basic GUI navigation skills and on higher-level reasoning about application states. The dataset is intended to simulate realistic end-user goals that CUAs may encounter, while avoiding tasks that require private user data or external configuration (e.g., importing files or logging into accounts).

\subsection{Autonomous Evaluation}

We propose a zero-shot method based on VLMs to automatically evaluate whether a task has been successfully completed\footnote{\url{https://github.com/martasumyk/vision-based-judge}}. Our pipeline consists of three main steps: 1. The CUA attempts to complete the given task; 2. A VLM receives the final screenshot along with the original task description and predicts whether the task has been successfully completed, providing a natural language justification for its decision; 3.  If the VLM judges the task as incomplete, its reasoning is fed back into the CUA. The agent then uses this feedback to attempt the task again, starting from its current state rather than restarting from scratch.

\begin{figure*}[t]
    \centering
    \includegraphics[width=\textwidth]{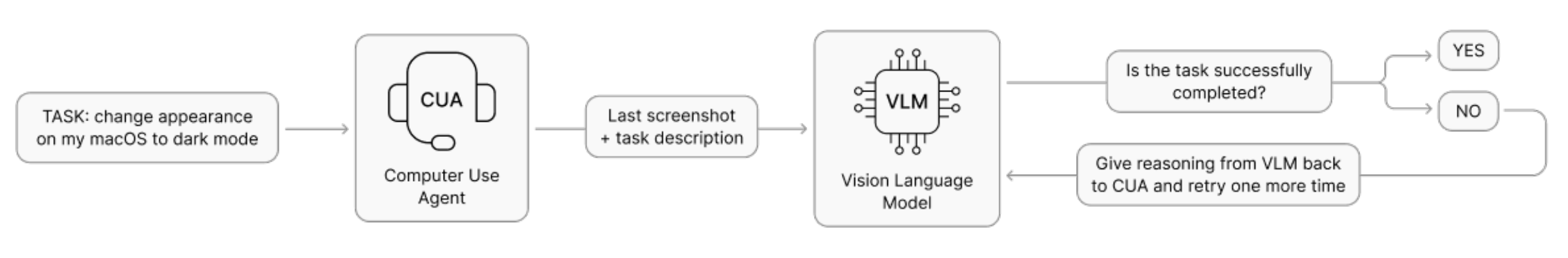}
    \caption{
        Overview of the proposed evaluation-feedback pipeline.
        CUA executes a user-defined task (e.g., \textit{“Change appearance on macOS to dark mode”}) and produces a final screenshot of the desktop state.
        The VLM receives the screenshot and task description, then judges whether the task was successfully completed.
        If the task is deemed incomplete, the VLM provides reasoning that is passed back to the CUA, which reattempts the task based on this feedback.
    }
    \label{fig:pipeline}
\end{figure*}

This feedback loop enables CUAs not only to receive automated success signals but also to adapt their behavior dynamically, reducing both task failure and redundant actions. The illustration of our proposed pipeline is shown in Figure~\ref{fig:pipeline}.

\subsubsection{Task Execution.}

For our experiments, we evaluated three CUAs: Claude Computer Use\footnote{\url{https://docs.claude.com/en/docs/agents-and-tools/tool-use/computer-use-tool}}, OpenAI Operator, and UI-TARS \cite{qin2025uitarspioneeringautomatedgui}. The first two are proprietary systems, while UI-TARS is open-source. We selected these agents because they currently achieve leading performance on the OSWorld benchmark \cite{xie2024osworldbenchmarkingmultimodalagents}.

Each CUA is provided with a task description and attempts to complete the task within the macOS environment. During execution, the full trajectory is recorded, including step-by-step screenshots, the actions performed at each step (where the action space consists of clicks, double-clicks, typing text, pressing keys, and waiting), and the agent's reasoning at each step.

\subsubsection{Outcome Evaluation.}

In this stage, the task description and the final screenshot of the desktop state are provided to a VLM, which is prompted in a zero-shot setting to determine whether the task has been successfully completed. The VLM produces both a binary judgment (\textit{done/not done}) and a short natural-language rationale explaining its decision. 

We evaluate five VLMs that represent both proprietary and open-source families. Among proprietary evaluators, we use GPT-4o\footnote{\url{https://openai.com/index/hello-gpt-4o/}} and Claude 3.5 Sonnet\footnote{\url{https://www.anthropic.com/claude}}, chosen for their state-of-the-art multimodal reasoning capabilities. For open-source models, we employ LLaVA-v1.5-7B \cite{liu2024llava15}, InternVL 2-8B \cite{chen2024internvl2}, and Qwen2-VL-7B \cite{bai2024qwen2vl}, which provide competitive performance. This selection covers a broad spectrum of parameter scales and training paradigms, allowing us to compare evaluation consistency across architectures and accessibility tiers.

This setup enables task evaluation to be performed independently of the acting CUA, reducing bias and ensuring that success is judged solely from observable interface states rather than internal model assumptions. Leveraging general-purpose vision–language reasoning allows the evaluator to robustly handle diverse applications and task types without requiring task-specific rules or heuristics.

\subsubsection{Feedback and Retry.}

If the VLM determines that the task has not been successfully completed, its rationale is passed back to the CUA as  feedback. The agent then uses this reasoning to replan and reattempt the task, resuming from its current state rather than restarting the entire trajectory. This feedback loop enables independent correction: the CUA can interpret evaluator feedback, and adjust its strategy accordingly. This feedback mechanism enables the agent to continue from its current state rather than restarting, leading to more efficient retries and higher overall task success rates.



\section{Results}

\subsection{Evaluator Accuracy Across CUAs}

The results in Table~\ref{tab:task_classification} show that accuracy of task completion classification, measured against human-annotated ground truth, is consistently high for both proprietary and open-source evaluators. Even in a zero-shot setting, most models demonstrate strong alignment with human judgments, confirming that vision-language models can reliably assess task success.

\begin{table*}[t]
\centering
\caption{
Task completion classification accuracy (\textit{done/not done}) across proprietary and open-source VLM-based evaluators for three CUAs. 
\textbf{Claude~3.5~Sonnet} achieves the highest proprietary performance, while \textbf{Qwen2-VL-7B} leads among open-source models.
}
\begin{tabular}{lccc}
\hline
\textbf{Evaluator Model} & \textbf{OpenAI Operator} & \textbf{Anthropic CU} & \textbf{UI-TARS} \\
\hline
\multicolumn{4}{c}{\textbf{Proprietary Evaluators}} \\
\hline
GPT-4o & 0.61 & 0.69 & 0.64 \\
Claude~3.5~Sonnet & \textbf{0.69} & \textbf{0.71} & \textbf{0.73} \\
\hline
\multicolumn{4}{c}{\textbf{Open-Source Evaluators}} \\
\hline
LLaVA-v1.5-7B & 0.56 & 0.61 & 0.52 \\
InternVL~2-8B & 0.62 & \textbf{0.67} & 0.61 \\
Qwen2-VL-7B & \textbf{0.68} & 0.66 & \textbf{0.70} \\
\hline
\end{tabular}
\label{tab:task_classification}
\end{table*}

\subsection{Effect of Evaluator Feedback on Success Rate}

\begin{figure}[t]
    \centering
    \includegraphics[width=\linewidth]{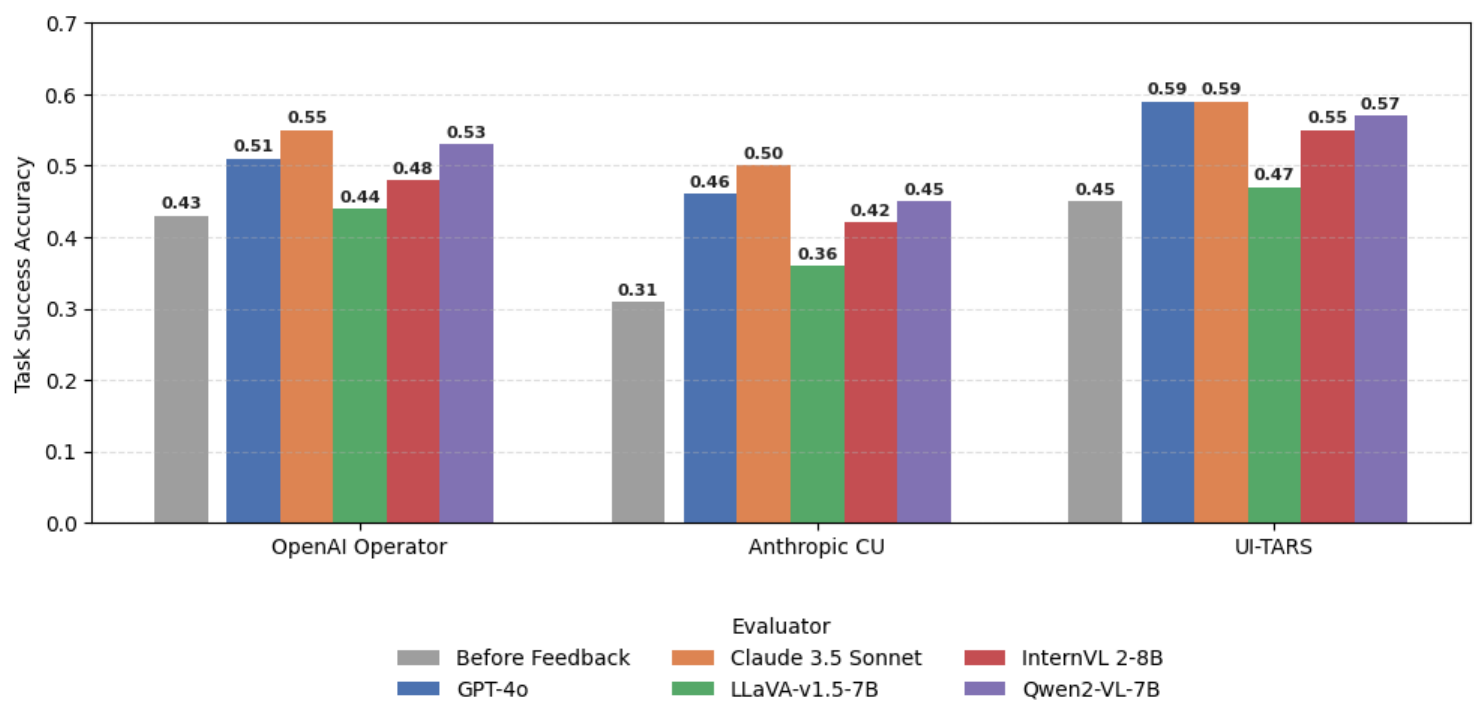}
    \caption{
Task success rates before and after evaluator feedback across three CUAs. 
Gray bars represent baseline success rates before feedback, while colored bars indicate post-feedback (after only one retry) success rate for five VLM evaluators. 
Proprietary evaluators (GPT-4o and Claude~3.5~Sonnet) achieve the largest relative gains, whereas open-source models (LLaVA-v1.5-7B, InternVL~2-8B, and Qwen2-VL-7B) provide consistent improvements across all agents.
}

    \label{fig:success_rate}
\end{figure}

Figure~\ref{fig:success_rate} illustrates the effect of evaluator feedback on task success rate across the three CUAs. All evaluated VLM feedback mechanisms lead to measurable performance gains compared to the baseline without feedback. Proprietary evaluators (GPT-4o and Claude~3.5~Sonnet) yield the largest improvements, achieving up to 61\% relative success rate gains, while open-source evaluators such as Qwen2-VL-7B also provide consistent boosts in success rate. Notably, agents with lower baseline success rate like Anthropic CU benefit the most from visual feedback, indicating that automated screen-based reasoning helps agents detect and correct incomplete actions. These findings highlight that VLM evaluators not only reliably assess task completion but also enhance the self-correction ability of CUAs through interpretable, vision-grounded feedback.

\section{Future Work}

Our task completion evaluation framework opens several promising directions for future research.

First, we plan to expand the framework beyond macOS to additional operating systems such as Linux and Windows.  Since fundamental interface components such as windows, menus, buttons, and text are common to Windows, Linux, and macOS, our vision-based approach eliminates the need for OS-specific instrumentation and enables straightforward transfer of the same evaluation logic to new environments. This cross-platform expansion would enable broader support of CUAs across all desktop environments.

Second, our current evaluation employs a binary success metric: a task is considered complete only when the final goal is reached. A natural extension is to develop step-level evaluation, in which each intermediate action is judged based on whether it moves the agent closer to the final objective. We also plan to conduct an ablation study to determine how many screenshots or temporal observations are most informative for reliable evaluation.

Third, we plan to conduct inter-model agreement analysis, calibration measurements, and consistency studies across different VLMs. These analyses will quantify how sensitive evaluations are to model choice and provide confidence intervals for task-success predictions incorporating techniques such as temperature scaling, conformal prediction, or ensemble averaging may further improve the robustness of judgments.

Forth, the evaluator’s output can be used directly as a reward signal within Reinforcement Learning (RL) pipelines for CUAs, providing interpretable, vision-grounded feedback that may improve exploration efficiency and stabilize long-horizon training. By replacing heuristic or human-provided rewards, this approach also reduces the need for large-scale human-labeled datasets in RL pipelines, enabling more scalable and autonomous agent training.

Finally, we aim to extend this work toward multi-agent frameworks \cite{zhuge2024agentasajudgeevaluateagentsagents, bhonsle2025autoevaljudgegeneralagentic}, where the evaluator continuously monitors CUA actions and delivers real-time feedback on each step. Such integration would allow agents to adapt their strategies dynamically, reduce redundant actions, and improve robustness. This could further enable the development of multi-agent systems in which specialized evaluators and actors collaborate to achieve complex computer-use goals.

\section{Conclusion}
We presented a framework that autonomously checks whether a CUA has completed its task using only the final screenshot and the task description. Instead of relying on hand-written scripts or system logs, our method uses VLMs to judge task success and provide short feedback that the agent can use to try the task completion again, resuming from the current state. We also provide a diverse, human-labeled dataset of $1{,}260$ tasks across $42$ built-in macOS applications to enable reproducible evaluation and support future research.

Across three CUAs and five VLM evaluators, our approach achieves up to \textbf{73\%} accuracy in identifying completed tasks and improves overall success rates by \textbf{27\%} on average. We find that weaker agents benefit the most, showing that external visual feedback can make CUAs more reliable and efficient.

Beyond improving accuracy, our framework provides a simple and general way to verify what agents actually achieve on screen. In the future, we plan to extend this work to other operating systems, explore step-by-step evaluation instead of only final results, and use the evaluator as a reward signal in reinforcement learning or multi-agent systems. This moves us closer to building CUAs that can not only act but also correctly recognize when their goals are accomplished.

\bibliography{aaai2026}

\end{document}